\documentclass{article}

\pdfoutput=1

\usepackage[numbers]{natbib}

\usepackage[utf8]{inputenc} 
\usepackage[T1]{fontenc}    
\usepackage{hyperref}       
\usepackage{url}            
\usepackage{booktabs}       
\usepackage{amsfonts}       
\usepackage{nicefrac}       
\usepackage{microtype}      
\usepackage{hyperref}
\hypersetup{
    colorlinks=true,
    linkcolor=blue,
    filecolor=magenta,      
    urlcolor=cyan,
    citecolor=red
}
\usepackage{bm}
\usepackage{tabu}
\usepackage{comment}
\usepackage{amsmath} 
\usepackage{subcaption}
\usepackage{listings} 
\usepackage{graphicx}
\usepackage{caption}
\usepackage{rotfloat}
\usepackage{float}
\usepackage{amsthm}
\usepackage{array}
\usepackage[linesnumbered, lined, ruled, boxed, commentsnumbered]{algorithm2e}
\graphicspath{{figures/}}

\begin{document}

\title{A comparison of methods for model selection when estimating individual treatment effects}

\author{Alejandro Schuler \\
	aschuler@stanford.edu \\
       Biomedical Informatics Research\\
       Stanford University\\
       Palo Alto, CA, USA 
          \and
     Michael Baiocchi \\
     baiocchi@stanford.edu \\
       Department of Medicine\\
       Stanford University\\
       Palo Alto, CA, USA 
       \and
            Robert Tibshirani \\
     tibs@stanford.edu \\
       Department of Statistics\\
       Stanford University\\
       Palo Alto, CA, USA 
	\and
     Nigam Shah \\
     nigam@stanford.edu \\
       Biomedical Informatics Research\\
       Stanford University\\
       Palo Alto, CA, USA }

\maketitle

\begin{abstract}
{Practitioners in medicine, business, political science, and other fields are increasingly aware that decisions should be personalized to each patient, customer, or voter. A given treatment (e.g. a drug or advertisement) should be administered only to those who will respond most positively, and certainly not to those who will be harmed by it. Individual-level treatment effects can be estimated with tools adapted from machine learning, but different models can yield contradictory estimates. Unlike risk prediction models, however, treatment effect models cannot be easily evaluated against each other using a held-out test set because the true treatment effect itself is never directly observed. Besides outcome prediction accuracy, several metrics that can leverage held-out data to evaluate treatment effects models have been proposed, but they are not widely used. We provide a didactic framework that elucidates the relationships between the different approaches and compare them all using a variety of simulations of both randomized and observational data. Our results show that researchers estimating heterogenous treatment effects need not limit themselves to a single model-fitting algorithm. Instead of relying on a single method, multiple models fit by a diverse set of algorithms should be evaluated against each other using an objective function learned from the validation set. The model minimizing that objective should be used for estimating the individual treatment effect for future individuals.}
\end{abstract} 

\section{Introduction}

The general decision problem we address is as follows: for a particular individual, a decision maker must choose between prescribing an intervention or no intervention. The intervention (treatment) may be a drug, an advertisement, a campaign email etc. The decision-maker's goal is to maximize some outcome for that patient or customer, which may be their lifespan, their net purchases, their political engagement etc. This is a causal inference problem because we seek to discover a causal relationship between the intervention and outcome. The causal \emph{treatment effect} is the difference between what would have happened had the individual been given the intervention and what would of happened had they not been. The outcomes under these different scenarios are referred to as \emph{potential outcomes} (\citealp{Rubin2005}).

Prior to the development of modern statistical methods, treatment policies were generally one-size-fits-all prescriptions based on estimates of the average treatment effect (\citealp{Segal:ub}). Experiments for inferring average effects limit individual heterogeneity by imposing strict criteria on the population under study (\citealp{Stuart:2014id}). Recently, however, researchers in multiple domains have attempted to leverage modern statistical technology and real-world data to tailor decisions to individuals; this phenomena is exemplified by the rise of personalized medicine (\citealp{Ferreira:2017fv}) and targeted advertisement (\citealp{Ascarza:2018ie, Matz:2017ix}). Decision-makers recognize that treating to the average, while expedient, does not result in the best outcome for all individuals (\citealp{Kravitz:2004fa,Segal:ub}). Attention is increasingly focused on the estimation of individual treatment effects.
\footnote{Note that estimating individual treatment effects is not the same as estimating personalized risks or prognoses with prediction models (e.g. for a heart attack, customer churn, or non-voting). Prediction models only predict what would happen to the individual given standard practice, not the difference of what would happen if a treatment were or were not given. As such, prediction models by themselves are often of little practical utility unless the effects of available treatments are known and relatively constant. If that is not the case, targeting treatment to individuals at high-risk for the outcome is not an optimal strategy: there may be high-risk individuals who do not respond or respond negatively to the treatment, and low-risk individuals who would respond very positively (\citealp{Ascarza:2018ie}).}

To clarify further discussion, we characterize each individual $i$ by a vector of pre-treatment features or covariates $x_i$, their treatment status (intervention $w_i=1$ or no intervention $w_i=0$) and their outcome $y_i$. Using the potential outcomes framework (\citealp{Rubin2005}), we write the potential outcomes under treatment and control as $Y(1)$ and $Y(0)$ and their conditional means as $\mu_0(x) = E[Y(0)|X=x]$ and $\mu_1(x) = E[Y(1)|X=x]$. The estimand in question is the conditional average treatment effect $\tau(x) = \mu_1(x) - \mu_0(x)$, which is the expected difference in potential outcomes under the alternative interventions for the individual in question. In different fields this quantity is alternatively called the individual treatment effect, individual causal effect, individual benefit, or individual lift. If the true conditional mean functions $\mu_w(x)$ are known, the rule (policy) that assigns each individual $x_i$ their optimal treatment is $d(x) = \underset{w \in \{0,1\}}{\text{argmax}} \ \ \mu_w(x)$, or, alternatively, $d(x) = I(\tau(x) > 0)$ where $I$ is the indicator function. Generally, the conditional mean functions $\mu_w$ are unknown, meaning that there is uncertainty about the individual treatment effect and optimal treatment policy.

There currently exist a number of methods to estimate individual treatment effects from randomized data. The process of estimating these effects is alternatively referred to as heterogenous effect modeling, uplift modeling, or individual treatment effect modeling. These approaches can also be used for observational data if certain assumptions are met or if combined with propensity score or matching techniques.

A manual subgroup analysis is the traditional approach to heterogenous treatment effect estimation. A subgroup analysis partitions the population of individuals into manually-specified subgroups and typically estimates an average treatment effect in each subgroup using traditional methods (i.e. linear or logistic regression) (\citealp{Gail:1985ft}). This approach requires a high degree of domain knowledge is prone to multiple-hypothesis testing problems if subgroups are not pre-specified. 

An alternative is to use any supervised learning method (e.g. LASSO, random forest, neural network) to fit functions $\hat\mu_0$ and $\hat\mu_1$ that estimate the conditional means $\mu_0$ and $\mu_1$ of the potential outcomes. These estimates are then used to estimate the treatment effect $\hat\tau(x) = \hat\mu_1(x) - \hat\mu_0(x)$ (\citealp{Gutierrez:2016tq, Austin:2012cy, Snowdn:2011ef}). This can be done by regressing the observed outcomes on the covariates in the untreated group to get $\hat\mu_0$ and regressing the observed outcomes on the covariates in the treated group to get $\hat\mu_1$. \citet{Kunzel:2017vg} call this approach T-learning (T for ``two models''). Similarly, it is possible to fit a single model $\hat\mu(x,w)$ and estimate the treatment effect in the same way as above by letting $\hat\mu_w(x) = \hat\mu(x,w)$ (S-learning, for single-model) \citealp{Kunzel:2017vg}. T-learning and S-learning together have been referred to as simulated twins, g-computation, counterfactual regression, or conditional mean regression. 

Modeling the conditional means is a valid approach, but many have noted that since the object of interest is the treatment effect we may be better off modeling it directly without appeal to the correctness of $\hat\mu_0$ and $\hat\mu_1$. Approaches in this vein include \citet{Zhao:2017vi}, \citet{Athey:2016wm}, \citet{Powers:2017wd}, and \citet{Nie:2017vi}. 

Among the variety of approaches and the number of hyper-parameter settings within each approach, which is best? As is the case with all statistical learning problems, there is no context-free way of knowing (\citealp{Wolpert:1996fp}): different methods will give better or worse estimates depending on the application. Indeed, using a large set of diverse simulations, \citet{Dorie:2017uo} find that the only somewhat consistent predictor of the success of a causal inference method is its ability to ``flexibly'' model the conditional means or treatment effect. Although that result surely depends upon the particulars of their simulations, it parallels the common knowledge in the machine learning literature that deep nets and additive regression trees often outperform linear models for real-world applications. However, even limiting ourselves to flexible treatment effect modeling methods, we are left with a panoply of approaches and hyper-parameter settings to chose from. 

We digress briefly to discuss the standard supervised learning setting where the task is to estimate $y$ given $x$ by building an estimator $\hat{\mu}(x)$. In this setting we can use the diversity of machine learning approaches to our advantage by performing model selection. Given $M$ modeling approaches and/or hyper-parameter settings, we build $M$ estimators $\{\hat \mu_1, \hat \mu_2 \dots \hat \mu_m \dots \hat \mu_M\}$. The quantity of interest in this case is the expected prediction risk of the model when it is applied to new data, according to some loss $l$. We express that as $E[l(\hat \mu_m(X), Y)]$. The idea of model selection is to estimate this expected risk for each of our models and find the model that minimizes it. There are several ways to estimate this risk, including information criteria, but data-splitting is the easiest and most widely applicable (\citealp{esl:2009wc, Arlot:2010fl, Dudoit:2005jw}). Before fitting the models, the observations are randomly split into training and validation samples. The models are fit on the training sample $\mathcal{T}$ and evaluated on the validation sample $\mathcal{V}$. The risk of each model is estimated with $\frac{1}{|\mathcal{V}|}\sum_{i \in \mathcal{V}}^{|\mathcal{V}|} l(\hat \mu_m (x_i), y_i)$. In cross-validation, this process is repeated round-robin across different random splits of the data and the estimated risks are averaged per model.

This approach breaks down for treatment effect estimation because the true treatment effect is never observed in any sample. In this case, the quantity of interest is the $\tau$-risk: $E[l(\hat\tau, \tau)]$. We would like to evaluate models $\{\hat\tau_1, \hat\tau_2 \dots \hat \tau_m \dots \hat \tau_M\}$ by estimating their $\tau$-risk on a validation set via $\frac{1}{|\mathcal{V}|}\sum_{i \in \mathcal{V}}^{|\mathcal{V}|}  l(\hat \tau_m (x_i), \tau(x_i))$ (this quantity has been called the precision in estimating heterogenous effects or PEHE by \citet{Hill2011}). The problem is that we never observe $\tau(x_i)$ directly (we only see one of the two potential outcomes) and thus have nothing to compare $\hat\tau(x_i)$ to. This estimator of $\tau$-risk is thus infeasible.

Several treatment effect model selection approaches have been suggested in the literature, but none of them enjoy the wide use and dominance that prediction error cross-validation has in the supervised learning setting. Most approaches are not general-purpose in that they require that the set of estimators $\{\hat\tau_1, \hat\tau_2 \dots \hat \tau_m \dots \hat \tau_M\}$ come from a specific class of models. For example, \citet{Powers:2017wd} and \citet{Athey2015} both use selection methods that are specific to the models they propose. The Focused Information Criterion (FIC) (\citealp{Claeskens:2003ck}) is a promising approach, but as of yet cannot be used to select between most machine learning estimators (\citealp{Jullum:2012uo}). \citet{Alaa:tj} propose an empirical Bayes approach for optimal prior selection.

It is clear that we lack a go-to general-purpose approach for applied researchers to select among treatment effect models, leaving the door open to poor practice. Using valid model selection will ensure that better models result from primary research, making it more certain, for instance, that a patient will benefit from their treatment or that an advertisement will reach interested parties. 

In this work, we summarize several model selection approaches that use an independent validation set to judge the quality of individual treatment effects estimated from a training set. Some of these approaches have been explicitly proposed for model selections- the rest are adapted from model fitting procedures proposed for individual treatment effect estimation or policy learning. We implement each of these approaches and test them against each other in a variety of experimental simulations where the usual assumptions of positivity, SUTVA, and ignorability hold. An outline of the paper is as follows. In section \ref{approaches}, we describe approaches for treatment effect model selection and introduce a didactic framework to relate them to each other. Section \ref{simulations} describes our experiments and results. We conclude in section \ref{conclusion} with a summary of our contributions and recommendations for researchers interested in estimating individual treatment effects.
 
\section{Metrics for treatment effect model selection via data-splitting}
\label{approaches}

As we have established, we are interested in statistics that, when optimized over a set of available model predictions on a validation set, also tend to minimize the $\tau$-risk $E[l(\hat\tau, \tau)]$. For the purposes of our discussion, we will focus on risk under squared error loss. Here we describe three approaches that fit the bill.

The first approach is to minimize the $\mu$-risk $E[l(\hat\mu_{W}(X), Y)]$. As we have seen, this quantity is easy to estimate. Furthermore, a perfect model of the potential outcomes implies a perfect model of the treatment effect, so we are justified in optimizing this quantity.

An alternative is to maximize the value of the treatment policy $\hat d(x) = I(\hat\tau(x) > 0)$ which indicates which individuals we expect to benefit from the treatment. The \emph{value} of a decision policy $\hat d$ is $v = E[Y|W=\hat d(X)]$. In other words, the value is the expected outcome of an individual when all individuals are treated according to the policy $\hat d$. If larger values of the outcome are more desirable (e.g. lifespan, click-through rate, approval ratings), then the policy that maximizes the value is optimal, and vice-versa. As we will see, this quantity is also simple to estimate in a few ways. If all we are interested is the treatment decision (and not the treatment effect itself), then we are already justified in maximizing value. However, $\hat\tau = \tau$ is a maximizer of the decision value $v$, so it may be justifiable to use the decision value to optimize for treatment effect estimation.

The last approach is to directly estimate the $\tau$-risk. We will see that there are several methods for doing so.

\subsection{$\mu$-risk}
\label{sec:pred-error}

There are many simple examples where minimizing the mean-squared error of predicted outcomes badly fails to select the model with the most accurate treatment effect \cite{Rolling:2013kz}. Despite this, we can attempt to use prediction error to select among treatment effect models. Assuming the treatment effect model is built by regressing the outcomes onto the covariates and treatment to obtain $\hat\mu_0$ and $\hat\mu_1$ (e.g.. with an S-learner or T-learner), we can estimate $\mu$-risk with

\begin{equation}
	\widehat{\mu\text{-risk}} = \frac{1}{|\mathcal{V}|} \sum_{i \in \mathcal{V}}^{|\mathcal{V}|}  
	(\hat \mu_{w_i} (x_i) - y_i)^2
\label{murisk}
\end{equation}
 
For individuals in the validation set who were treated ($w=1$), we estimate their outcome using the treated model and assess error, and vice-versa for the untreated. This is equivalent to estimating the predictive risk separately for $\hat\mu_1$ and $\hat\mu_0$. 

If treatment assignment is random conditional on observed covariates, we can appropriately weight each residual with the estimated inverse propensity of observing an individual $x_i$ with treatment $w_i$ as suggested by \citet{VanderLaan:2003ir}

\begin{equation}
	\widehat{\mu\text{-risk}}_{IPTW} = \frac{1}{|\mathcal{V}|} \sum_{i \in \mathcal{V}}^{|\mathcal{V}|}  
	\frac{(\hat \mu_{w_i} (x_i) - y_i)^2}{\check p_{w_i}(x_i)}
\label{murisk-iptw}
\end{equation}

Where $p_{w_i}(x_i) = P(W=w_i | X=x_i)$. This is the propensity score if $w_i = 1$ and one minus the propensity score if $w_i = 0$. The notation $\check a$ indicates a quantity that is estimated using only data in the validation set, whereas $\hat a$ is estimated using only data in the training set. The effect of this is to create the correct ``pseduo-population'' that would have been observed under random assignment. I.e. if a treated individual $x_i$ had a probability of $0.1$ of being assigned the treatment under the observed nonrandom assignment, their residual should be weighted by a factor of $10$ to account for the 9 other individuals who would have received that treatment had the assignment been random.

\subsection{Value}
\label{sec:value}

\citet{Kapelner:3baXYEjR} and \citet{Zhao:2017wa} propose the same validation set estimator for the value of a treatment effect model:

\begin{equation}
\label{value}
\hat v_{IPTW} = \frac{1}{|\mathcal{V}|}\sum_{\mathcal{V}} \frac{y_i I(w_i=\hat d(x_i))}{\check p_{w_i}(x_i)}
\end{equation}

where again $\check p_{w_i}(x_i)$ is an estimate of $P(W=w_i | X=x_i)$.  We call this the inverse propensity of treatment weighted (IPTW) value estimator.\footnote{
We should note that this estimator is closely related to one commonly used in the direct marketing literature called uplift:

\begin{equation}
\label{gain-basic}
	\frac{1}{|\mathcal V |} \left(
		  \frac{\sum_{\mathcal{V}} y_i  \hat d(x_i) w_i}{\sum_{\mathcal{V}}  \hat d(x_i) w_i} - 
		  \frac{\sum_{\mathcal{V}} y_i  \hat d(x_i) (1-w_i)}{\sum_{\mathcal{V}}  \hat d(x_i)  (1-w_i)} 
		  \right)
		  \sum_{\mathcal{V}} \hat d(x_i) 
\end{equation}

This is actually a special case of

\begin{equation}
\label{gain}
	\hat g  = \dfrac{1}{|\mathcal V |} \left[ \sum_{\mathcal{V}} \dfrac{y_i  \hat d(x_i) w_i}{p(x_i)} - \sum_{\mathcal{V}} \dfrac{y_i  \hat d(x_i) (1-w_i)}{1-p(x_i)} \right]
\end{equation}

To see this, we rewrite equation \ref{gain-basic} as:
\[
	 \dfrac{1}{|\mathcal V |} \underbrace{ \frac{ \sum_{\mathcal{V}} \hat d(x_i)}{\sum_{\mathcal{V}}  \hat d(x_i) w_i} }_{1/\hat p}
		  	\sum_{\mathcal{V}} y_i  \hat d(x_i) w_i - 
		\dfrac{1}{|\mathcal V |}  \underbrace{ \frac{ \sum_{\mathcal{V}} \hat d(x_i)}{\sum_{\mathcal{V}}  \hat d(x_i)  (1-w_i)} }_{1/ (1-\hat p)}
		  	\sum_{\mathcal{V}} y_i  \hat d(x_i) (1-w_i) 
\]

The multipliers underbraced above are unbiased estimates of $1/ p_{w_i}$ because of the conditional independence of $\hat d(X)$ and $W$. Thus the traditional estimator (equation \ref{gain-basic}) is suitable for use when the propensity score $p(x) = p$ is a constant, as is the case in randomized experiments.

To see the relationship between uplift and value, note that
\[
\begin{array}{rcl}	
	E[\hat g | X=x] & = & E[\tau(x) \hat d(x) | X=x]  \\
	& =  & (\mu_1(x)-\mu_0(x))  \hat d(x)  \\
	E[\hat v | X=x] & = & E[Y | W=\hat d(x), X=x]  \\
	& = & \mu_0(x)(1-\hat d(x)) + \mu_1(x)\hat d(x)
\end{array}
\]

Consider two policies $\hat d_a$ and $\hat d_b$ and their respective estimated values $\hat v_a, \hat v_b$  and gains $\hat g_a, \hat g_b$. The expected difference in value between the two models is 

\[
\begin{array}{rcl}
E[\hat v_a - \hat v_b] 

& = & E[E[\hat v_a | X] - E[\hat v_b|X]] \\

&=& E \left[
	\mu_0(X)(1-\hat d_a(X)) + \mu_1(X)\hat d_a(X) 
      - \mu_0(X)(1-\hat d_b(X)) -  \mu_1(X)\hat d_b(X)
\right] \\

&=& E \left[
	  \hat d_a(X) (\mu_1(X)  - \mu_0(X) ) 
	- \hat d_b(X) (\mu_1(X)  - \mu_0(X) )
\right] \\

&=& E \left[
	  E[\hat g_a | X] ) 
	- E[\hat g_b | X] )
\right] \\

&=& E \left[ \hat g_a - \hat g_b \right] \\
\end{array}
\]

Thus a model optimizing any unbiased estimate of uplift also optimizes any unbiased estimate of value in expectation.
}

In the randomized setting where $p_{w_i}(x_i) = 0.5$, we can imagine that two side-by-side experiments were run, one in which treatments were assigned according to the model ($W = \hat d(X)$) and one in which they were assigned according to the opposite recommendation ($W = 1 - \hat d(X)$). The data in the validation set are a concatenation of the data from these two experiments. To estimate the value of our model, we average the outcomes of individuals in the first experiment and ignore the data from the second experiment. This is essentially what the estimator in equation \ref{value} is doing. When $p_{w_i}(x_i) \ne 0.5$, we must appropriately weigh the outcomes according to the probability of treatment to accomplish the same goal. \citet{Kapelner:3baXYEjR} give a similar explanation, but omit the role of the propensity score. \citet{Zhao:2017wa} provide a short proof that $\hat v$ is unbiased for the true value $v = E[Y|W = \hat d(X)]$. 

A problem with this estimator is that it depends on the correctness of the propensity model. In addition, it only utilizes a portion of the data: $\hat v_{IPTW}$ throws away individuals whose treatments do not match $\hat d(x_i)$. In the spirit of \citet{Dudik:tn}, \citet{Athey:wj} overcome this by using a doubly-robust formulation

\begin{equation}
	\hat v_{DR} = \frac{1}{|\mathcal{V}|}\sum_{\mathcal{V}}
	\hat d(x_i)
	\left(
	\check\mu_1(x_i) - \check\mu_0(x_i)+ (2w_i -1)\frac{y_i - \check\mu_{w_i}(x_i)}{\check p_{w_i}(x_i)}
	\right)
\label{value-dr}
\end{equation}

where $\check\mu_{w_i}(x_i)$ and $\check p(x_i)$ can be estimated with standard regression methods using data in the validation set. \citet{Athey:wj} use an estimator of this form in order to fit a policy model and establish theoretical gaurentees, whereas here we will use it to select among several pre-fit models.

\subsection{$\tau$-risk}

We have already seen that $\frac{1}{|\mathcal{V}|}\sum_{i \in \mathcal{V}}^{|\mathcal{V}|}  (\hat \tau_m (x_i)- \tau(x_i))^2$ is infeasible because $\tau(x_i)$ is never observed directly. A natural workaround is to replace $\tau(x_i)$ with an estimate derived from the validation set:

\[
\frac{1}{|\mathcal{V}|}\sum_{i \in \mathcal{V}}^{|\mathcal{V}|}  (\hat \tau (x_i) -  \check \tau_i)^2
\]

Here, $\check \tau$ is a plug-in estimate of $\tau$ estimated using data in the validation set $\mathcal{V}$. 

\citet{Rolling:2013kz} propose an estimator $\check \tau_i$ based on matched treated and control individuals in the validation set. Briefly, for each individual $i$ in the validation set they use Mahalanobis distance matching to identify the most similar individual $\bar{i}$ in the validation set with the opposite treatment ($w_i \ne w_{\bar i}$) and compute $\check \tau_i = (2w_i -1)(y_i - y_{\bar i})$ as the plug-in estimate of $\tau(x_i)$. 

\begin{equation}
\widehat{\tau\text{-risk}}_{match} = \frac{1}{|\mathcal{V}|}\sum_{i \in \mathcal{V}}^{|\mathcal{V}|}  (\hat \tau (x_i) - (2w_i -1)(y_i - y_{\bar i}))^2
\label{trisk-match}
\end{equation}

They prove under general assumptions and a squared-error loss that a more mathematically tractable version of their algorithm has selection consistency, meaning that it correctly selects the best model as the number of individuals goes to infinity. They conjecture that the practical version of the algorithm retains this property.

A downside of this approach are that Mahalanobis matching scales relatively poorly and matches become difficult to find in high-dimensional covariate spaces. An alternative proposed by \citet{Gutierrez:2016tq} takes advantage of the fact that the IPTW-weighted (transformed) outcome $\frac{(2W-1)Y}{p_W(X)}$ is an estimator for $\tau$:

\begin{equation}
\widehat{\tau\text{-risk}}_{IPTW} = 
	\frac{1}{|\mathcal{V}|}\sum_{i \in \mathcal{V}}^{|\mathcal{V}|}  
	\left(\hat \tau (x_i) - \frac{(2w_i -1)y_i}{\check p_{w_i}(x_i)}\right)^2
\label{trisk-iptw}
\end{equation}

This formulation is also used for model fitting in the transformed-outcome forest of \citet{Powers:2017wd} and in some versions of the causal tree in \citet{Athey2015}.

Our final approach deviates from this schema and is due to \citet{Nie:2017vi}, who propose minimizing

\begin{equation}
\widehat{\tau\text{-risk}}_{R} = 
	\frac{1}{|\mathcal{V}|}\sum_{i \in \mathcal{V}}^{|\mathcal{V}|}  
	((y_i - \check m(x_i)) - (w_i - \check p(x_i))\hat\tau (x_i))^2
\label{trisk-r}
\end{equation}

The function $\check m(x)$ is an estimate of $E[Y|X]$ which can be obtained by regressing $Y$ onto $X$ without using the treatment $W$. \citet{Nie:2017vi} provide theoretical and empirical results that show how fitting models using an objective of this form (with some additional stipulations) can outperform T- and S-learning. We propose using this same construction to select among models fit by arbitrary means.

\section{Experiments}
\label{simulations}

\subsection{Overview}

We demonstrate the utility of these approaches using simulations. Each simulation is defined by a data-generating process with a known effect function, which allows us to compute true test set errors. Each run of each simulation generates a dataset, which we split into training, validation, and test samples. We use the training data to estimate $M$ different treatment effect functions $\hat\tau_m$ using $M$ different algorithms (e.g. S-learning with gradient boosted trees). For each of those we calculate each validation metric (e.g. $\widehat{\mu\text{-risk}}$, $\hat v_{IPTW}$, $\widehat{\tau\text{-risk}}_{R}$) using the data in validation set. The models selected by optimizing each validation-set metric are then used to estimate treatment effects on the test set. The test-set treatment effect estimates of each model are compared to the known effects to calculate the true cost of using each metric for model selection. Each simulation is repeated multiple times. All of the code used to set up, run, and analyze the simulations is feely available on \href{https://github.com/som-shahlab/ITE-model-selection}{github}.

\subsubsection{Data-generating processes and sampling}

We use the sixteen simulations from \citet{Powers:2017wd}, each of which we repeat $100$ times. The data-generating process in each simulation all satisfy the usual assumptions made about ignorability, SUTVA, and positivity, but vary in the amount of nonlinearity in the conditional mean functions and signal-to-noise ratio. Details may be found in \citet{Powers:2017wd}. 

In each repetition of each simulation, $1000$ samples are used for training ($\mathcal{T}$), $1000$ for validation ($\mathcal{V}$), and $1000$ for testing ($\mathcal{S}$). Let the data be denoted by $\{\bm y, \bm w, \bm x\} = \{[y_1 \dots y_n]^T, [w_1 \dots w_n]^T, [x_1 \dots x_n]^T\}$ where $n = |\mathcal{T}| + |\mathcal{V} | + |\mathcal{S}|$. Let $\bm y^{(\mathcal A)}$ represent $[y_i | i \in \mathcal A]$. Let the set of treated individuals be $\mathcal W_1 = \{i | w_i =1\}$ and let the set of untreated individuals be $\mathcal W_0 = \{i | w_i =0\}$. In a slight abuse of notation, let $f(\bm x) = [f(x_i) \dots f(x_n)]$.

\subsubsection{Learning Algorithms}

For each simulated dataset, we use the training sample to estimate $M$ different treatment effect functions $\hat\tau_m$ using $M$ different algorithms. The learning algorithms we use here are defined by a unique combination of a meta-learning approach (S-, T-, or R-) \cite{Kunzel:2017vg, Nie:2017vi}, a learning algorithm (e.g. elastic net or gradient boosted trees), and a set of hyperparameter values for that learning algorithm. This provides us with a broad set of estimated functions to select among. It would also be possible to use other estimators (e.g. causal forests or causal boosting) with varying hyperparameter values to estimate individual treatment effect functions, but in this work we restrict ourselves to meta-learning approaches because they can be used off-the-shelf to accommodate a large variety of machine learning algorithms.

\paragraph{S-learners}  In the S-learning framework, a single model $\hat\mu(x,w)$ is fit by regressing $\bm y^{(\mathcal T)}$ onto $[\bm x^{(\mathcal T)}, \bm z^{(\mathcal T)}]$ to produce $\hat\mu(x,w)$. The term $\bm z^{(\mathcal T)}$ is determined by $z_i = ( w_i- 0.5) x_i$ and ensures that the treatment effect is not implicitly regularized \cite{Nie:2017vi}. The treatment effect is calculated as $\hat\tau(x) = \hat\mu(x,1) - \hat\mu(x,0)$. 

The learning algorithms we use to fit $\hat\mu$ are gradient boosted trees (number of trees ranging from $1$ to $500$, tree depth of $3$, shrinkage of $0.2$ and minimum $3$ individuals per node) and elastic nets ($\alpha=0.5$, $\lambda \in [e^{-5}, e^2]$). These models give us a range of high-performing linear and nonlinear models to select among. We estimate a treatment effect model $\hat\tau_m(x)$ for each combination of algorithm and hyperparameters.

\paragraph{T-learners} A T-learner is fit by separately regressing $\bm y^{(\mathcal T \cap \mathcal W_0)}$ onto $\bm x^{(\mathcal T \cap \mathcal W_0)}$ to estimate $\hat\mu_0(x)$ and $\bm y^{(\mathcal T \cap \mathcal W_1)}$ onto $\bm x^{(\mathcal T \cap \mathcal W_1)}$ to estimate $\hat\mu_1(x)$. The treatment effect is calculated as $\hat\tau(x) = \hat\mu_1(x) - \hat\mu_0(x)$. 

We use the same algorithms and hyperparameters as above to fit the models $\hat\mu_w(x)$. When estimating the various treatment effect models $\hat\tau_m(x)$, we only consider combinations of models $\hat\mu_1$ and $\hat\mu_0$ that were fit using the same algorithm with the same hyperparameters. This increases computational efficiency, but need not be done in practice (i.e. $\hat\mu_0$ could be fit using a linear model and $\hat\mu_1$ fit using a random forest). It is an appropriate simplification to make in our experiments since our focus is on model selection and not model fitting.

\paragraph{R-learners} In the R-learning framework of \citet{Nie:2017vi}, treatment effect models are estimated by minimizing 

\begin{equation}
	\frac{1}{|\mathcal{T}|}\sum_{i \in \mathcal{T}}^{|\mathcal{T}|}  
	((y_i - \hat m(x_i)) - (w_i - \hat p(x_i))t(x_i))^2
\label{r-learner}
\end{equation}

over a space of candidate models $t(x)$. Given $\hat m(x)$ and $\hat p(x)$, this is equivalent to a weighted least-squares problem with a pseudo-outcome of $\frac{ y_i - \hat m(x_i) }{w_i - \hat p(x_i)}$ and weights of $(w_i - \hat p(x_i))^2$. As such, it can be solved using a variety of learning algorithms. 

In our experiments, we estimate the quantities $\hat m(\bm x^{(\mathcal T)})$ and $\hat p(\bm x^{(\mathcal T)})$ once per simulated dataset using cross-validated cross-estimation over the training set \cite{Nie:2017vi, Wager:2016dz}. The internal cross-validation is run over estimates derived from the same combinations of algorithms and hyperpameters as used by the S- and T-learners. These estimates are then fixed and used for all R-learners. Each R-learner $\hat\tau_m(x)$ is produced by minimizing equation \ref{r-learner}, again using each combination $m$ of learning algorithm and hyperparameter values.

\subsubsection{Model selection metrics}

The following metrics are used to select among models in each simulation:


\begin{center}
\tabulinesep=1mm
\begin{tabu}{|l|c|l|}
	\hline
	 \rowfont[c]{\bfseries} Metric & Equation &  Reference \\
	 \hline
	 Random 						& NA  			& NA \\
	 $\widehat{\mu\text{-risk}}$		& \ref{murisk}  		& NA \\
	 $\widehat{\mu\text{-risk}}_{IPTW}$	& \ref{murisk-iptw} 	& \citet{VanderLaan:2003ir} \\
	 $\hat v_{IPTW}$ 				& \ref{value}  		& \citet{Zhao:2017wa} \\
	 $\hat v_{DR}$					& \ref{value-dr}		& \citet{Athey:wj}  \\
	 $\widehat{\tau\text{-risk}}_{match}$	& \ref{trisk-match}	& \citet{Rolling:2013kz} \\
	 $\widehat{\tau\text{-risk}}_{IPTW}$ 	& \ref{trisk-iptw} 	& \citet{Gutierrez:2016tq} \\
	 $\widehat{\tau\text{-risk}}_{R}$		& \ref{trisk-r} 		& \citet{Nie:2017vi} \\
	 \hline
\end{tabu}
\label{metric-table}
\end{center}

Taking the model that minimizes (or maximizes, when appropriate) one of these metrics defines a model selection approach. 

The ``random'' approach selects a model uniformly at random from the available models. 

Several of these metrics require estimates $\check p(\bm x^{(\mathcal V)})$, $\check \mu_w(\bm x^{(\mathcal V)})$, or $\check m(\bm x^{(\mathcal V)})$. In our experiments, we estimate each of these quantities using cross-validated cross-estimation over the validation set alone. The internal cross-validation is run over estimates derived from the same combinations of algorithms and hyperpameters as used by the treatment effect learners. 

\subsubsection{Evaluation metrics}

Let the model $\hat\tau_m$ selected by optimizing metric $h$ be written as $\hat\tau^{*_h}$. 

We are interested in the quantities

\[
\tau\text{-risk}_h = E[ (\hat\tau^{*_h} (X) - \tau(X))^2 ]
\]

and 

\[
v_h = E[ Y| W =\hat d^{*_h} (X)]
\]

which we unbiasedly estimate in a large test set $\mathcal{S}$ via

\begin{equation}
\label{true-mse}
\tau\text{-risk}^{(\mathcal{S})}_h = \frac{1}{|\mathcal{S}|}\sum_{i \in \mathcal{S}} (\hat\tau^{*_h} (x_i) - \tau(x_i))^2
\end{equation}

and 

\begin{equation}
\label{true-value}
v^{(\mathcal{S})}_h = \frac{1}{|\mathcal{S}|}\sum_{i \in \mathcal{S}} \mu_{\hat d^{*_h}(x_i)}(x_i)
\end{equation}

Where $\hat d^{*_h}(x_i) = I(\hat\tau^{*_h}(x_i) > 0)$ as before. 

$\tau\text{-risk}_{h}^{(\mathcal{S})}$ calculates how well the selected model estimates the treatment effect for individuals in the test set. $v^{(\mathcal{S})}$ is the decision value of applying the treatment policy $\hat d(x)$ derived from each selected model to the individuals in the test set.

These are both useful metrics, although only the first ($\tau\text{-risk}$), sometimes called ``precision in estimating heteorgenous effects'', or PEHE) has typically been used in simulation studies in the individual effect estimation literature while $v^{(\mathcal{S})}$ is used in the policy learning literature. To see why they are both important, consider two models ($A$ and $B$) that estimate the same treatment effect for all individuals, except for two individuals ($i=1$ and $i=2$). Let $\tau(x_1) = \tau(x_2) = 0.1$, i.e. both individuals would benefit from the treatment in reality. Model $A$ estimates $\hat\tau_A(x_1) = -0.1$ and $\hat\tau_A(x_2) = 0.1$. In other words, it incorrectly suggests not treating individual $1$, although the absolute difference $|\hat\tau_A(x_1)-\tau(x_1)| = 0.2$ is quite small, so it is not heavily penalized according to $\tau\text{-risk}$. Model $B$ estimates $\hat\tau_B(x_1) = 0.1$ and $\hat\tau_B(x_2) = 100$. Model $B$ correctly suggests the treatment for both individuals, but the absolute difference $|\hat\tau_B(x_2)-\tau(x_2)| = 99.9$ is large and is heavily penalized by $\tau\text{-risk}$. Often, what we want is a model that correctly assigns treatment to the individuals who stand to benefit from it. Using $\tau\text{-risk}$ in this case would favor model $A$ even though it leads to the mistreatment of more individuals than model $B$ does. However, $\tau\text{-risk}$ is still a useful metric. There may be cases where a researcher is interested in the precise magnitude of the effect for each individual, perhaps so that scarce resources can be allocated most effectively. 

\begin{figure}
  \centering
    \includegraphics[width=0.6\textwidth]{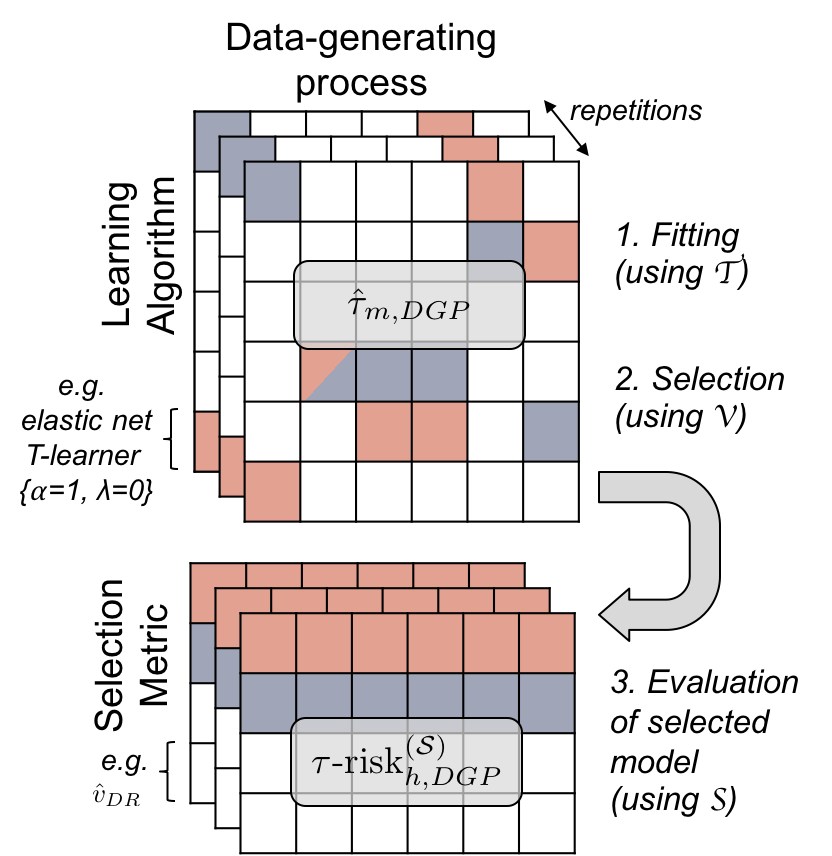}
      \caption{A visual representation of our experiments. Each row in the top grid represents one learning algorithm. Each column is one of the data-generating processes (DGPs). Each cell in that grid thus represents a fit function $\hat\tau_{m,DGP}(x)$ obtained by using that meta-learner, learning algorithm, and hyperparameter values on a training sample $\mathcal T$ from that DGP. The colored cells represent the models selected by optimizing each metric in table \ref{metric-table} using the validation set $\mathcal V$. Different colors represent different the different metrics. The models chosen via each metric in each DGP are gathered together (bottom grid) and used to estimate the individual treatment effect on a test-set $\mathcal S$ which we use to assess the true generalization error $\tau\text{-risk}^{(\mathcal S)}$ (and policy value $v^{(\mathcal S)}$) attained by using each model selection metric. The entire process is repeated and the reported results are averages across repetitions.}
      \label{experiment-fig}
\end{figure}

\subsection{Results}

\begin{table}[ht]
\centering
\caption{$\tau\text{-risk}^{(\mathcal{S})}$. True test-set $\tau\text{-risk}$ of the treatment effect model selected according to each metric (rows) in each simulation (columns), averaged across replications. Simulations 9-16 use the same data-generating functions as 1-8, but with nonrandom treatment assignment. Smaller numbers indicate better performance.}
\begin{tabular}{lllllllll}
  \hline
metric & 1 & 2 & 3 & 4 & 5 & 6 & 7 & 8 \\ 
  \hline
Random & 0.102 & 2.684 & 2.223 & 1.718 & 1.233 & 2.845 & 5.859 & 2.677 \\ 
  $\widehat{\mu\text{-risk}}$ & \bfseries{0.007} & 0.235 & \bfseries{0.014} & 0.808 & 0.212 & 0.963 & 2.970 & 1.738 \\ 
  $\widehat{\mu\text{-risk}}_{IPTW}$ & \bfseries{0.007} & 0.235 & \bfseries{0.014} & 0.808 & 0.212 & 0.963 & 2.970 & 1.738 \\ 
  $\hat v_{IPTW}$ & 0.130 & 0.634 & 4.278 & 0.690 & 0.738 & 0.887 & 4.300 & 1.513 \\ 
  $\hat v_{DR}$ & 0.052 & 0.969 & 8.458 & 0.785 & 4.466 & 0.972 & 5.759 & 2.577 \\ 
  $\widehat{\tau\text{-risk}}_{match}$ & 0.013 & 0.542 & 0.964 & 0.068 & 0.260 & 0.970 & 3.681 & 0.815 \\ 
  $\widehat{\tau\text{-risk}}_{IPTW}$ & 0.056 & 0.101 & 0.050 & 0.154 & 0.288 & \bfseries{0.786} & 2.913 & 0.746 \\ 
  $\widehat{\tau\text{-risk}}_{R}$ & 0.008 & \bfseries{0.059} & 0.017 & \bfseries{0.035} & \bfseries{0.043} & \bfseries{0.786} & \bfseries{2.794} & \bfseries{0.736} \\ 
  \hline
   \\
  \\
  \hline
metric & 9 & 10 & 11 & 12 & 13 & 14 & 15 & 16 \\ 
  \hline
Random & 0.401 & 3.178 & 61.830 & 4.143 & 4.173 & 3.019 & 8.899 & 4.235 \\ 
  $\widehat{\mu\text{-risk}}$ & 0.012 & 0.290 & \bfseries{0.020} & 0.797 & \bfseries{0.055} & 1.277 & \bfseries{3.900} & 2.495 \\ 
  $\widehat{\mu\text{-risk}}_{IPTW}$ & 0.013 & \bfseries{0.288} & 0.040 & 0.782 & 0.060 & 1.250 & 4.050 & 2.277 \\ 
  $\hat v_{IPTW}$ & 0.822 & 1.290 & 4.542 & 1.677 & 5.247 & 1.274 & 6.174 & 2.431 \\ 
  $\hat v_{DR}$ & 0.146 & 1.169 & 5.678 & 1.821 & 26.571 & 1.356 & 7.340 & 2.668 \\ 
  $\widehat{\tau\text{-risk}}_{match}$ & 0.400 & 1.617 & 4.400 & 1.769 & 9.399 & 1.425 & 4.284 & 1.661 \\ 
  $\widehat{\tau\text{-risk}}_{IPTW}$ & 1.049 & 0.451 & 2.387 & 1.129 & 4.260 & \bfseries{1.120} & 4.133 & 2.123 \\ 
  $\widehat{\tau\text{-risk}}_{R}$ & \bfseries{0.008} & 0.348 & 1.986 & \bfseries{0.149} & 0.078 & 1.169 & 3.923 & \bfseries{1.539} \\ 
     \hline
\end{tabular}
\label{tmse-table}
\end{table}

\begin{table}[h!t]
\centering
\caption{$v^{(\mathcal{S})}$. True test-set value of the decision rule derived from the treatment effect model selected according to each metric (rows) in each simulation (columns), averaged across replications. Simulations 9-16 use the same data-generating functions as 1-8, but with nonrandom treatment assignment. Larger numbers indicate better performance.}
\begin{tabular}{lllllllll}
  \hline
metric & 1 & 2 & 3 & 4 & 5 & 6 & 7 & 8 \\ 
  \hline
Random & \bfseries{0.008} & 0.501 & 1.914 & 0.791 & 0.771 & 0.446 & 0.138 & 0.685 \\ 
  $\widehat{\mu\text{-risk}}$ & \bfseries{0.008} & 0.795 & \bfseries{1.998} & 0.966 & 0.906 & 0.663 & 0.642 & 0.835 \\ 
  $\widehat{\mu\text{-risk}}_{IPTW}$ & \bfseries{0.008} & 0.795 & \bfseries{1.998} & 0.966 & 0.906 & 0.663 & 0.642 & 0.835 \\ 
  $\hat v_{IPTW}$ & \bfseries{0.008} & 0.798 & 1.989 & 0.969 & 0.877 & \bfseries{0.678} & 0.651 & 0.842 \\ 
  $\hat v_{DR}$ & \bfseries{0.008} & 0.770 & 1.978 & 0.969 & 0.486 & 0.667 & 0.354 & 0.784 \\ 
  $\widehat{\tau\text{-risk}}_{match}$ & \bfseries{0.008} & 0.783 & 1.998 & 0.984 & 0.903 & 0.671 & 0.505 & 0.881 \\ 
  $\widehat{\tau\text{-risk}}_{IPTW}$ & \bfseries{0.008} & 0.806 & 1.997 & 0.979 & 0.890 & 0.678 & 0.672 & \bfseries{0.884} \\ 
  $\widehat{\tau\text{-risk}}_{R}$ & \bfseries{0.008} & \bfseries{0.809} & 1.998 & \bfseries{0.986} & \bfseries{0.907} & 0.678 & \bfseries{0.687} & 0.883 \\ 
  \hline
   \\
  \\
  \hline
metric & 9 & 10 & 11 & 12 & 13 & 14 & 15 & 16 \\ 
  \hline
Random & \bfseries{-0.011} & 0.407 & 1.852 & 0.505 & 0.680 & 0.421 & 0.157 & 0.586 \\ 
  $\widehat{\mu\text{-risk}}$ & \bfseries{-0.011} & 0.762 & \bfseries{1.993} & 0.945 & \bfseries{0.906} & 0.653 & 0.532 & 0.709 \\ 
  $\widehat{\mu\text{-risk}}_{IPTW}$ & \bfseries{-0.011} & 0.761 & 1.993 & 0.945 & 0.906 & 0.653 & 0.505 & 0.725 \\ 
  $\hat v_{IPTW}$ & \bfseries{-0.011} & 0.721 & 1.923 & 0.881 & 0.562 & \bfseries{0.656} & \bfseries{0.593} & 0.751 \\ 
  $\hat v_{DR}$ & \bfseries{-0.011} & 0.703 & 1.969 & 0.945 & 0.142 & 0.654 & 0.177 & 0.671 \\ 
  $\widehat{\tau\text{-risk}}_{match}$ & \bfseries{-0.011} & 0.494 & 1.960 & 0.861 & 0.126 & 0.647 & 0.481 & 0.790 \\ 
  $\widehat{\tau\text{-risk}}_{IPTW}$ & \bfseries{-0.011} & 0.740 & 1.966 & 0.921 & 0.598 & 0.653 & 0.579 & 0.786 \\ 
  $\widehat{\tau\text{-risk}}_{R}$ & \bfseries{-0.011} & \bfseries{0.777} & 1.983 & \bfseries{0.995} & 0.904 & 0.655 & 0.547 & \bfseries{0.806} \\ 
   \hline
\end{tabular}
\label{value-table}
\end{table}

\begin{figure}
  \centering
    \includegraphics[width=0.9\textwidth]{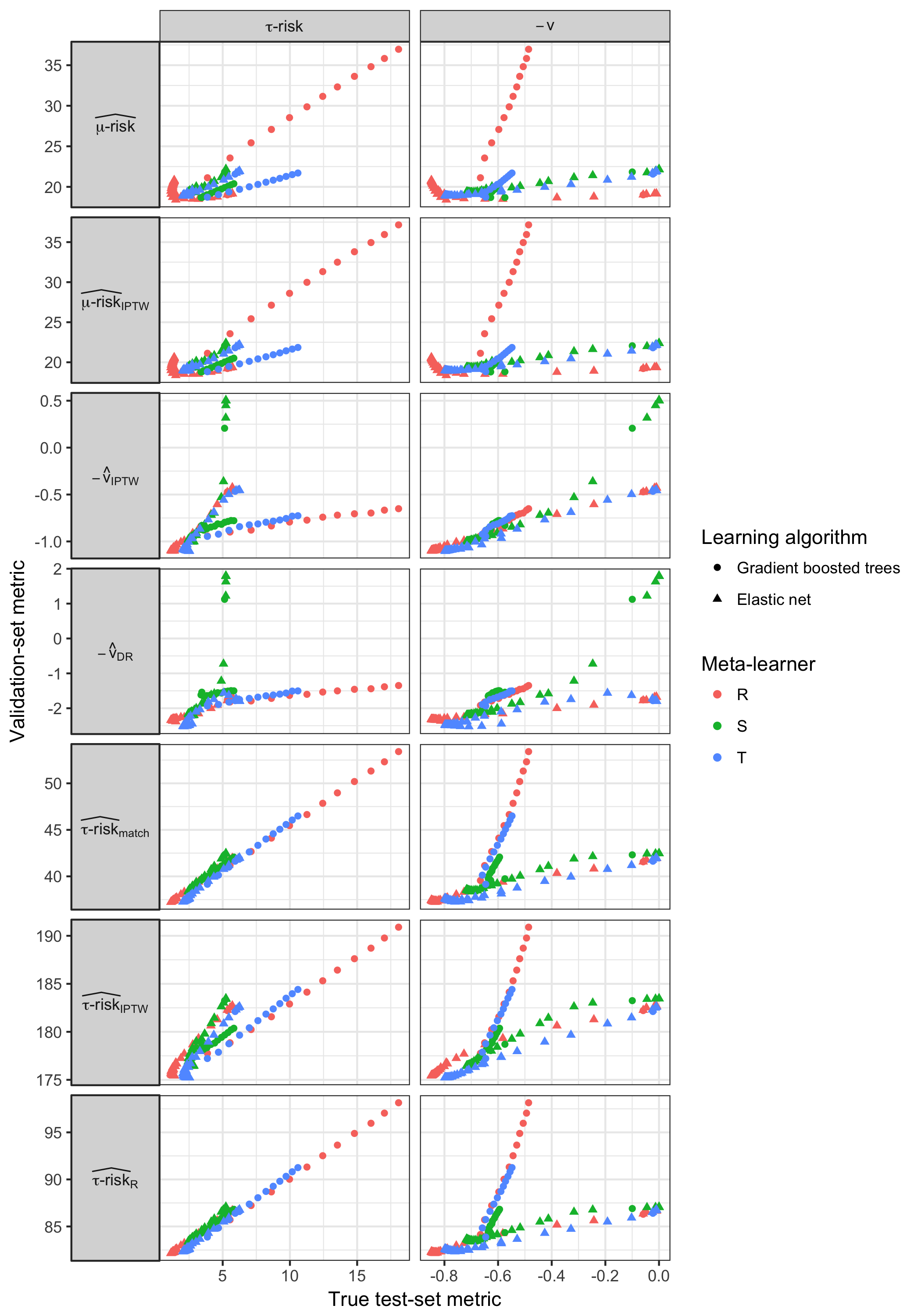}
      \caption{Validation-set and test-set metrics for all treatment effect models, averaged over all replications of simulation 16 (equivalent plots for other simulations show similar patterns). Each point is a single model resulting from a unique combination of meta-learner (S, T, or R), learning algorithm, and hyperparameters. The x-axis facets are the values of $\tau\text{-risk}^{(\mathcal{S})}$ and $-v^{(\mathcal{S})}$ for each model (lower values indicating higher performance). The y-axis facets are the model selection metrics for each model as estimated on the validation set (lower values indicating higher predicted performance). High correlation between a validation-set and test-set metric indicates good performance of the model selection metric.}
      \label{details16}
\end{figure}

The results in tables \ref{tmse-table} and \ref{value-table} show that $\widehat{\tau\text{-risk}}_{R}$ is the metric that most consistently selects models with low $\tau\text{-risk}$ and high value. This is especially true when treatment assignment is randomized, but also to a large extent when the assignment is biased. In the simulations where other model selection approaches win out, it is typically by a small margin. Both $\widehat{\mu\text{-risk}}$ and $\widehat{\mu\text{-risk}}_{IPTW}$ perform well (the two are identical in randomized settings). Even when the goal is to optimize value, selecting on the basis of these metrics leads to better performance than using $\hat v_{IPTW}$ or $\hat v_{DR}$. Scenarios 1 and 9 have zero treatment effect for all individuals, which is why columns 1 and 9 in table \ref{value-table} are identical: all policies will lead to the same outcomes for all patients. Figure \ref{details16} shows the results from simulation 16 in greater detail.

\section{Conclusion}
\label{conclusion}

Although prediction error cross-validation is widely used to select between predictive models, there is no consensus on how to perform model selection for individual treatment effects models. We use simulations to examine the performance of several proposed approaches and our own adaptations of methods that have been previously used for model fitting, but not model selection. 

Our results show that $\widehat{\tau\text{-risk}}_{R}$ is the validation set metric that, when optimized, most consistently leads to the selection of a high-performing model. This conclusion strengthens the claims of \citet{Nie:2017vi} and extends the utility of their framework to a model selection setting where treatment effects models may be fit by any algorithm. 

Figure \ref{details16} shows that all of the estimators of $\tau$-risk are biased upwards (i.e. they overestimate the risk). An explanation of that phenomenon for $\widehat{\tau\text{-risk}}_{IPTW}$ can be found in \citet{Gutierrez:2016tq}. Their argument extends to $\widehat{\tau\text{-risk}}_{match}$ if the matching treatment effect estimate $(2w_i -1)(y_i - y_{\bar i})$ is unbiased. Regardless, for the purposes of model selection, the relevant quantity is the difference in estimates for different models and so this bias is of less importance. Figure \ref{details16} shows good correlation between $\tau$-risk estimated on the validation set according to various methods and the true test-set $\tau$-risk for each model. 

Interestingly, our results also show that model selection on the basis of $\hat v_{IPTW}$ or $\hat v_{DR}$ is generally suboptimal, even when the goal is to optimize $v^{(\mathcal S)}$. This is likely because these metrics are in a sense ``coarser'' than the other metrics we consider, effectively turning a regression problem into a classification problem. However, figure \ref{details16} does show that $\hat v_{IPTW}$ and $\hat v_{DR}$ have low empirical bias for $v^{(\mathcal S)}$, as they should when standard assumptions of unconfoundedness are met. Thus, unlike estimators of $\tau$-risk, it is appropriate to use these estimators to gauge the value of the final treatment decision policy created by thresholding a treatment effect model.

The primary limitation of our work is that it relies on simulations. This is a necessity because $\tau\text{-risk}_{h}^{(\mathcal{S})}$ and $v^{\mathcal{(S)}}$ cannot be calculated from real data. While no set of simulations can mimic the full range of real-world data-generating processes, our experiments span a range of linear and nonlinear potential outcome functions, high- and low-noise settings, and levels of bias in treatment assignment. 

To conclude, we advocate for the use of $\widehat{\tau\text{-risk}}_{R}$ as a model selection metric for individual treatment effects models. Researchers estimating heterogenous treatment effects need not limit themselves to a single model-fitting algorithm. There are many cases in our simulations where models fit via an R-learner are outperformed by models fit via a T-learner, or where using the elastic net confers an advantage over using gradient boosted trees, and vice-versa. Instead of relying on a single method, multiple models fit by a diverse set of algorithms should be evaluated against each other using $\widehat{\tau\text{-risk}}_{R}$ as estimated on a validation set. The best performing model should be used for estimating the individual treatment effect for future individuals.


\bibliographystyle{plainnat}
\bibliography{references}

\end{document}